\documentclass[nojss, noheadings, nofooter]{jss}

\usepackage{tabto}
\usepackage{amsmath}
\usepackage{amsfonts}
\usepackage{arydshln}
\usepackage{etoolbox}
\usepackage{orcidlink}

\title{
\pkg{FairLangProc}: A \proglang{Python} package for fairness in NLP
}
\author{
    Arturo Pérez-Peralta~\orcidlink{0009-0007-1613-0634}\\
    {\scriptsize Universidad Carlos III de Madrid}
    \And 
    Sandra Benítez-Peña~\orcidlink{0000-0002-6246-2847}\\
    {\scriptsize Department of Statistics} \\ 
    {\scriptsize uc3m-Santander Big Data Institute}\\
    {\scriptsize Universidad Carlos III de Madrid} 
    \And 
    Rosa E. Lillo~\orcidlink{0000-0003-0802-4691}\\
    {\scriptsize Department of Statistics} \\
     {\scriptsize uc3m-Santander Big Data Institute}\\ 
    {\scriptsize Universidad Carlos III de Madrid} 
}

\Plainauthor{Sandra Benítez-Peña, Second Author} %%
\Plainauthor{Rosa E. Lillo, Second Author} %% comma-separated
\Plaintitle{FairLangProc: A Python package for fairness in NLP} %% without formatting
\Shorttitle{\pkg{FairLangProc}: A \proglang{Python} package for fairness in NLP} %% a short title (if necessary)

\Abstract{
  The rise in usage of Large Language Models to near ubiquitousness in recent years has risen societal concern about their applications in decision-making contexts, such as organizational justice or healthcare. This, in turn, poses questions about the fairness of these models in critical settings, which leads to the developement of different procedures to address bias in Natural Language Processing. Although many datasets, metrics and algorithms have been proposed to measure and mitigate harmful prejudice in Natural Language Processing, their implementation is diverse and far from centralized. As a response, this paper presents \pkg{FairLangProc}, a comprehensive \proglang{Python} package providing a common implementation of some of the more recent advances in fairness in Natural Language Processing providing an interface compatible with the famous Hugging Face \pkg{transformers} library, aiming to encourage the widespread use and democratization of bias mitigation techniques. The implementation can be found on \url{https://github.com/arturo-perez-peralta/FairLangProc}.
}
\Plainkeywords{keywords, comma-separated, not capitalized, Java} %% without formatting
\Address{
  Arturo Pérez-Peralta\\
  Department of Statistics\\
  Universidad Carlos III de Madrid\\
  Madrid, Spain\\
  ORCID: 0009-0007-1613-0634\\
  E-mail: \email{100507525@alumnos.uc3m.es}\\
  URL: \url{http://eeecon.uibk.ac.at/~zeileis/}
}
\begin{document}

%% include your article here, just as usual
%% Note that you should use the \pkg{}, \proglang{} and \code{} commands.

\section{Introduction}

The astonishing results of the transformer architecture on Natural Language Processing (NLP) tasks \citep{BERT, GPT}, their scalation properties \citep{vaswani2017attention} and the massive amount of text data available \citep{wang2019superglue, commoncrawl} have led to the development of Large Language Models (LLM) whose performance towers above that of traditional Language Models (LM) \citep{CPM-LLM1, BLOOM-LLM2}. Furthermore, LLMs have been widely adopted for custom downstream tasks by leveraging the flexibility provided by fine-tuning \citep{FINETUNING} and their few-shot learning capabilities \citep{FEWSHOT}, establishing a new zeitgeist in the NLP community. These factors have led to their widespread adoption across major areas of society such as academia \citep{ACADEMIA, ACADEMIA2}; industry,  including sectors such as finance \citep{FINANCE}, healthcare \citep{HEALTH} or law \citep{LAW} and personal use, for example, as a personal assistant or search engine \citep{SEARCHENGINE, MICROSOFT}. Furthermore, the recent surge in their reasoning ability \citep{COT} and the development of cost-efficient models \citep{DEEPSEEK} suggest that there are still new avenues for improvement.

However, this paradigm shift opens the possibility of societal harm through the perpetuation of existing biases in the training data \citep{PERPETUATE}. This concern is at the center of the field of algorithmic fairness, whose community has paid close attention to the propagation of prejudices against disadvantaged groups through Machine Learning (ML), measuring discrimination in critical contexts such as finance, healthcare, and organizational justice \citep{INDUSTRY}. These biases can only be mitigated through a proactive approach, otherwise risking unfair treatment of sensitive groups \citep{FAIRMLBOOK}. The existing methods for debiasing can be classified according to their position on the ML pipeline, distinguishing between pre-processors, who curate the training dataset; in-processors, which modify the learning procedure; and post-processors, which affect the outputs of a given model \citep{SURVEYMLFAIRNESS}.

A myriad of methods have been proposed to debias LLMs, ranging from simply rewriting their outputs \citep{REWRITING} to projecting the hidden representations of the model to a bias-free subspace \citep{SENT}. However, although numerous algorithms have been proposed, their practical implementations are frequently inaccessible, with many never being released publicly, which poses a challenge to practitioners in both academia and industry.

To address this problem, this paper presents \pkg{FairLangProc}, a \proglang{Python} package that provides a common implementation of existing fairness datasets, metrics and processors, encouraging the democratization of these tools in NLP. This work broadly follows \cite{SURVEY}, revisiting the literature of bias mitigation in LLMs and showing how to use the implementation of each method. Concretely, our contributions consists of an interface that permits easy handling of datasets for bias evaluation and an implementation of a myriad of fairness metrics and a plethora of debiasing pre-processors, in-processors and intra-processors. Furthermore, this paper provides a comprehensive review of the implemented method, delving into the theoretical background of these tools and showcasing the underpinnings of each module by explaining their parameters and showing usage examples. If this was not enough, the company repository provides a series of notebooks with all the details of the different examples, and the documentation page of the package is equipped with in-depth explanations of all the methods. Finally, we also performed a case study which serves as a rough comparison of the performance and fairness different debiasing processors.

The paper is structured as follows. In Section~\ref{sec:LMandFairness} we provide a quick tour of the core concepts, the notation that will be used, and the installation of the package. The next three Sections will explain the main features of the package, with Section~\ref{sec:Datasets} delving on datasets, Section~\ref{sec:metrics} explaining the different metrics that have been implemented and Section~\ref{sec:processors} going into detail on the different bias mitigation strategies. The exposition will alternate theoretical explanations of the chosen methods with code chunks that showcase the use of the package in a real-world setting. To see how the different blocks fit together, Section~\ref{sec:CaseStudy} features a more complete use case with numerical results applying most pieces of the package in a single experiment, showcasing the power of the implementation. Finally, Section~\ref{sec:conclusions} concludes with a recapitulation of the most important points and functionalities of the library.

\section{Language models and fairness}\label{sec:LMandFairness}

This Section introduces both the notation that will be used when debiasing Language Models and the underlying theoretical framework, followed by a first contact with \pkg{FairLangProc} explaining its context and how to install it.

Starting with the basics, text corpora composed of multiple words, sentences, or texts will be denoted using blackboard-bold symbols, $\mathbb{S}$. Individual sentences or texts will be represented by uppercase letters, $S$, while lowercase letters, $w$, will denote individual words. Finally, the hidden representation vectors will be denoted by boldface letters, $\mathbf{h}$. For simplicity, assume that the input text has already been tokenized. Therefore, the input space may be represented by $\mathbb{R}^L$ where $L$ is the maximum sequence length. In this context, a LM, $M$, generates a vector representation for the sequences, although sometimes we can also consider an intermediate layer of said model. Formally, let $M: \mathbb{R}^{L} \longrightarrow \mathbb{R}^{d_{m}}$  where and $d_{m}$ is the latent dimension of the model. Given an input sequence $S$, its embedding or hidden representation is denoted by $\mathbf{h} = M(S)$. Note that this terminology encompasses encoder-only, decoder-only and encoder-decoder architectures \citep{NLPBOOK}.

The hidden representation vector is semantically rich, allowing its use in downstream tasks. For this purpose, a head, $g:\mathbb{R}^{d_{m}} \longrightarrow \mathbb{R}^{d_o}$, where $d_o$ is the number of outputs, must be trained later. The models are usually trained for classification; that is, each observation $S$ is given a label $y$, which the model predictions, $\hat{y}  = g(M(S))$, must match. The model parameters are then optimized through gradient descent by specifying a certain loss function, $\mathcal{L}$, which generally takes the form of the cross-entropy \citep{NLPBOOK},

\begin{equation*}
    \mathcal{L} = -\sum_{k} y_k \log {\hat{y}_k},
\end{equation*}

where the sum is taken over the training instances. Of particular interest is the case of language modeling, in which the labels are tokens whose probabilities must be assigned.

Finally, sensitive information such as race, gender identity or religion will be represented through a discrete variable, $A$, signifying the different social groups arising from it. Following the previous notation, blackboard-bold, $\mathbb{A}$, will denote corpora of text containing sensitive information, either at the word or sentence level, and lowercase, $a$, individual sensitive words or sentences.

To address the problem of assessing and mitigating bias in LMs we have developed \pkg{FairLangProc}, a \proglang{Python} package incorporating datasets, metrics, and methods that aim to measure, identify and handle harmful societal prejudice in NLP tasks. To install our package, simply run:

\begin{CodeInput}
>>> pip install FairLangProc
\end{CodeInput}

The package depends only on mainstream open-source libraries, the minimum tested versions being:

\begin{itemize}
    \item \proglang{Python} $\geq 3.10$
    \item \pkg{numpy} $\geq 2.2.4$
    \item \pkg{pandas} $\geq 2.2.3$
    \item \pkg{scikit-learn} $\geq 1.6.1$
    \item \pkg{torch} $\geq 2.6.0$
    \item \pkg{transformers} $\geq 4.47.1$
    \item \pkg{datasets} $\geq 3.4.1$
    \item \pkg{adapter-transformers} $\geq 1.1.0$
    \item \pkg{pytest} $\geq 8.4.1$
\end{itemize}

To make sure everything is set up correctly, there is a \code{tests} folder in the package's repository with a myriad of tests that check the correctness of the implementation. All tests can be run with the following terminal line:

\begin{CodeInput}
python -m pytest -v
\end{CodeInput}

Finally, \pkg{FairLangProg} comprehensive documentation\footnote{\url{https://fairlangproc.readthedocs.io/en/latest/}} with explanations of all the methods and classes implemented and their parameters, providing usage examples and detailed exposition of their theoretical background.

\section{Fairness datasets}\label{sec:Datasets}

The NLP community has created and compiled multiple datasets for bias evaluation with many different methods and metrics. We will not delve into this topic, those interested in a comprehensive list and description of existing datasets are refered to \cite{SURVEY}, where they introduce a taxonomy based on their structure, distinguishing between those based on counterfactual inputs and those based on generated text from prompts. In \pkg{FairLangProc} these datasets have to be downloaded from the \code{Fair-LLM-Benchmark}\footnote{\url{https://github.com/i-gallegos/Fair-LLM-Benchmark}} repository, which should be cloned inside the \code{datasets} folder. Once this is done, the datasets can be accessed through the  \code{BiasDataLoader} method with the following parameters:

\begin{itemize}
    \item \textbf{dataset}: \code{str}. Name of the dataset. When doing \code{dataset = None} a list of available datasets is printed, which can be found in Table \ref{tab:datasets}.
    \item \textbf{config}: \code{str}. This parameter is used to choose among concrete instances of a given dataset if there are different versions. For example, imagine a dataset that allows for prejudice measurement across different societal groups (distinguishing between age, gender, race,...). The user could then choose which version they are interested with the \code{config} parameter. When setting \code{config = None}, a list of all possible values will be printed.
    \item \textbf{format}: \code{str}. Controls the output type. The available formats are \code{hf} for a Hugging Face \pkg{dataset} dataset, \code{pt} for a \pkg{Pytorch} dataset, and \code{raw} for raw dictionaries containing pandas data frames or string lists.
\end{itemize}

Here is an usage example:

\begin{CodeInput}
>>> from FairLangProc.datasets import BiasDataLoader
>>> data = BiasDataLoader(dataset = "BUG", config = "gold", format = "hf")
\end{CodeInput}

This piece of code stores the gold version of the BUG dataset \citep{BUG} inside the \code{data} variable as a Hugging Face dataset.

\begin{table}[h]
    \centering
    \begin{tabular}{|c|c|c|}
    \hline
        Dataset & Size & Reference \\
        \hline
        \hline
        BBQ & $58,492$ & \cite{BBQ} \\
        \hline
        BEC-Pro & $5,400$  & \cite{BEC} \\
        \hline
        BOLD & $23,679$ & \cite{BOLD} \\
        \hline
        BUG & $108,419$ & \cite{BUG} \\
        \hline
        Crow-SPairs & $1,508$ & \cite{CPS} \\
        \hline
        GAP & $8,908$ & \cite{GAP} \\
        \hline
        HolisticBias & $460,000$ & \cite{Holistic} \\
        \hline
        HONEST & $420$ &  \cite{HONEST} \\
        \hline
        StereoSet & $16,995$ & \cite{STEREO}\\
        \hline
        UnQover & $30$ & \cite{UnQover} \\
        \hline
        WinoBias+ & $1,367$ &\cite{REWRITING} \\
        \hline
        WinoBias & $3,160$ & \cite{WinoBias} \\
        \hline
        WinoGender  & $720$ & \cite{WinoGender} \\
        \hline

%        \hline
%        GAP-Subjective & $8,908$ & Gender & & \cite{GAPSub}\\
%
%        \hline
%        WinoQueer & $45,540$ & Sexual orientation & & \cite{WinoQueer}\\
%        \hline
%        RedditBias & $11,873$ & Gender, race, religion, sexual orientation & & \cite{Reddit}\\
%        \hline
%        Bias-STS-B & $16,980$ & Gender & & \cite{BiasSTSB} \\
%        \hline
%        PANDA & $98,583$ & Age, gender,race & & \cite{PANDA} \\
%        \hline
%        Equity Evaluation Corpus & $4,320$ & Gender, race & & \cite{EquityEval} \\
%        \hline
%        Bias NLI & $5,712,066$ & Gender, nationality, religion & & \cite{BiasNLI} \\
%        \hline
%        RealToxicityPrompts & $100,000$ & Masked Language modeling & & \cite{RealToxicity} \\
%
%        \hline
%        TrustGPT & $9$ & Masked Language modeling & & \cite{TrustGPT} \\
%
%        \hline
%        Gre-BiasIR & $118$ & Masked Language modeling & & \cite{GBIR} \\
          
    \end{tabular}
    \caption{Available datasets for bias evaluation}
    \label{tab:datasets}
\end{table}

\section{Fairness metrics}\label{sec:metrics}

This section introduces the different fairness metrics that haven been implemented in the library to measure discrimination in NLP. Broadly, they can be classified into three categories:

\begin{enumerate}
    \item Embedding metrics: if they measure bias by examining the model's hidden representations of input text.
    \item Probability metrics: if they measure bias by computing the probabilities of certain tokens or sentences.
    \item Generated text metrics: if they measure bias by examining text generated by the model, looking for harmful or stereotypical words.
\end{enumerate}

In \pkg{FairLangProc} they can be accessed through the \code{metrics} submodule.

\subsection{Embedding metrics}

Embedding metrics aim to measure bias in the model's hidden representation vectors ope\-ra\-ting under the assumption that a set of similar words or sentences which differ only on their demographic information should be close in latent space. The most famous embedding metric is given by \cite{WEAT} in the Word Embedding Association Test (WEAT), which aims to measure associations between demographic and neutral attributes. Demographic attributes are usually binary and denoted by $A_1, A_2$, signifying two different societal groups (male and female, christians and atheist,...) and represented by a corpora of words $\mathbb{A}_1$ and $\mathbb{A}_2$ (he, him, son and she, her, daughter; priest, nun, pope and non-believer, heretic, sceptic,...). Neutral attributes, on the other hand, are denoted by $W_1, W_2$ and represent two different stereotypes whose demographic association we are interested in. These stereotypes are likewise associated with a corresponding corpora of words, $\mathbb{W}_1, \mathbb{W}_2$ ranging from the occupational (technical and care work, proffesions and home roles,...) or academic (mathematics and arts, engineering and social sciences, medicine and nursing,...) to harm and prejudice (competence and incompetence, insults and praises,..), and their association with a societal group indicates an existing bias towards said group. Embedding metrics measure this association through the cosine similarity of the embeddings of words belonging to text corpora associated with each neutral attribute, $\mathbb{W}_i$,

\begin{equation*}
    s(a, W_1, W_2) = \sum_{w_1\in \mathbb{W}_1} \frac{
\cos(a, w_1)}{|\mathbb{W}_1|} - \sum_{w_2\in \mathbb{W}_2} \frac{
\cos(a, w_2)}{|\mathbb{W}_2|},
    \label{WEATSIM}
\end{equation*}

where $a$ represents the embedding of an arbitrary word and $s$ represents its similarity to the neutral attributes, with a positive score signifying an association with $W_1$ while a negative score implies a correlation with $W_2$. In principle the WEAT test can measure the association of any two concepts $A_1$ and $A_2$ to the attributes $W_1$ and $W_2$. In their original paper, \cite{WEAT} run WEAT on $10$ different sets of target words and attributes, resulting in $10$ tests, not all of them concerned with societal bias. For example, WEAT $1$ measures the association of bugs and flowers with pleasant and unpleasant words, which has no relation to prejudice; while WEAT $3$ measures the correlation between European American and African American names to pleasant and unpleasant words, which has a clear implication on group bias. Nonetheless, we will usually focus on the case where $A_1$ and $A_2$ represent sensitive demographic groups.

In any case, WEAT then measures bias through the effect size, which computes the average similarity between a corpora of text related to each sensitive attribute, $\mathbb{A}_i$, and the neutral parameters,

\begin{equation*}
    WEAT(A_1, A_2, W_1, W_2) = \frac{\sum_{a_1 \in \mathbb{A}_1} s(a_1, W_1, W_2)/ |\mathbb{A}_1| - \sum_{a_2 \in \mathbb{A}_2} s(a_2, W_1, w_2)/ |\mathbb{A}_2| }{\text{std}_{a\in \mathbb{A}_1 \cup \mathbb{A}_2} s(a, W_1, W_2)}.
    \label{WEATTEST}
\end{equation*}

A large effect size in either direction indicates a strong bias at the semantic level. This test can be run at the word \citep{WEAT} or sentence \citep{SEAT} level, and it can be further generalized to contextualized embeddings \citep{CEAT}. These metrics are implemented through the \code{WEAT} abstract class that can be found in the \code{metrics} submodule. The most relevant methods of said class are the following:

\begin{itemize}
    \item \code{WEAT.__init__(self, model, tokenizer, device)}: Initialization of the class. Stores the encoder, the tokenizer and the device where the model will be ran.
    \item \code{WEAT.metric(self, W1_words, W2_words, A1_words, A2_words, n_perm, pval)}:\\
    Main method to run the association test. The \code{word} parameters are lists of words which play the role of $W_1, W_2, A_1, A_2$. \code{pval} is a boolean variable which runs the permutation test to determine the statistical significance of the measured significance. If \code{pval} is set to true, \code{n_perm} is an intenger that represents the number of permutations of the test.
    \item \code{WEAT._get_embedding(self)}: Abstract method. The user should specify how the embeddings of the inputs should be extracted from the output of the model.
\end{itemize}

Here we show an usage example of how to measure association scores in BERT \citep{BERT}:

\begin{CodeInput}
>>> from FairLangProc.metrics import WEAT

>>> class BertWEAT(WEAT):
...     def _get_embedding(self, outputs):
...         return outputs.last_hidden_state[:, 0, :]
>>> tokenizer = AutoTokenizer.from_pretrained('bert-base-uncased')
>>> model = AutoModel.from_pretrained('bert-base-uncased')
>>> weatClass = BertWEAT(model = model, tokenizer = tokenizer)

# Measure the association of math and art words to binary gender identity
>>> math = ['math', 'algebra', 'geometry', 'calculus', 'equations'] 
>>> arts = ['poetry', 'art', 'dance', 'literature', 'novel']
>>> masc = ['male', 'man', 'boy', 'brother', 'he']
>>> femn = ['female', 'woman', 'girl', 'sister', 'she']

>>> weatClass.metric(
    W1_words = math, W2_words = arts,
    A1_words = masc, A2_words = femn,
    pval = False
    )
\end{CodeInput}

\subsection{Probability metrics}

Probability metrics rely on the computation of the probability of masked tokens through a language model head. A mask is a special token which is used during the training stage of Language Models to hide another semantically meaningful token. The model is then assigned the task of guessing which token hides behind the mask using the context of the rest of the text or just what came before. In any case, probability metrics can be classified into two categories depending on whether they rely on the probabilities of a single token of the sentence (masked token methods) or if they mask the whole sentence word by word (pseudo loglikelihood methods).

\subsubsection{Masked token metrics}

These methods rely on compute the probability of individual masked tokens. The main metrics in this regard are the Log-Probability Bias Score (LPBS) \citep{LPBS} and the Categorical Bias Score (CBS) \citep{CBS}. Broadly, they rely on predicting the probability that certain words with demographic connotation appear in the "[X]" spot in sentences of the form "[X] is [Y]". For example, if "[Y]" is "an engineer" the template now becomes "[X] is an engineer". We can now use a Language Model to compute the probabilities that words with demographic information appear in the "[X]" spot to measure the bias the Language Model has toward associating these demographics with engineering. Following with the example, if we used "he" and "she" and computed the probabilities of "he is an engineer" and "she is an engineer", we might obtain that the masculine version of the sentence is more likely, infering a bias manifested in the association of masculine words and engineering. Moreover, if the feminine version was more likely we could infer a bias in the opposite direction.

On the one hand, LPBS measures bias for a binary demographic group. It computes the predicted probability for a token $a$, $p_a$, using the template "[MASK] is [NEUTRAL ATTRIBUTE]"; and normalizes it by computing the model's prior probability, $p_{a, prior}$, based on the template "[MASK] is [MASK]". The score is computed as the difference of the logarithms of the normalized probabilities,

\begin{equation*}
    \text{LPBS} = \log\frac{p_1}{p_{prior, 1}} - \log\frac{p_2}{p_{prior, 2}}.
\end{equation*}

On the other hand, CBS provides a generalization of LPBS for non-binary demographic attributes like religion or race by using the variance over all the protected attribute words, $\mathbb{A}$:

\begin{equation*}
    \text{CBS} =  \text{Var}_{a\in \mathbb{A}}\log\frac{p_a}{p_{prior, a}}.
\end{equation*}

Their implementation is provided by the \code{LPBS}, \code{CBS} methods in the \code{metrics} submodule. Their use is nearly identical: they use very similar parameters, the only difference being if the demographic words are given in pairs or $n-$tuples. The full list of parameters is given by:
\begin{itemize}
    \item \textbf{model}, \code{torch.nn.Module}: Model used to compute the probabilities. 
    \item \textbf{tokenizer}, \code{TokenizerType}: Tokenizer associated with the model. 
    \item \textbf{sentences}, \code{list[str]}: List of masked sentences.
    \item \textbf{target\_words}, \code{list[tuple[str]]}: List of words with demographic information. When using \code{LPBS} the tuples should store pairs of words, while \code{CBS} allows for tuples with multiple words.
    \item \textbf{fill\_words}, \code{list[str]}: List of words of neutral attributes.
    \item \textbf{mask\_indices}, \code{list[int]}: List of integers (either $0$ or $1$) which represent the position of the mask of the target word ($0$ if it is the first mask, $1$ if it is the second one). 
\end{itemize}

A small example can help clarifying their usage:

\begin{CodeInput}
>>> from transformers import AutoTokenizer, AutoModelForMaskedLM
>>> from FairLangProc.metrics import LPBS, CBS

>>> sentences = [
...     "[MASK] is a [MASK].",
...     "[MASK] is a [MASK].",
... "The [MASK] was a [MASK]."
... ]
>>> target_words = [
...     ("John", "Mary"),
...     ("He", "She"),
...     ("man", "woman")
... ]
>>> fill_words = [
...     "engineer",
...     "nurse",
...     "doctor"
... ]
>>> mask_indices = [0, 0, 1]
>>> model = AutoModelForMaskedLM.from_pretrained('bert-base-uncased')
>>> tokenizer = AutoTokenizer.from_pretrained('bert-base-uncased')

>>> LPBSscore = LPBS(
...     model = model,
...     tokenizer = tokenizer,
...     sentences = sentences,
...     target_words = target_words,
...     fill_words = fill_words,
...     mask_indices = mask_indices
... )

>>> target_words = [
...     ("John", "Mamadouk", "Liu"),
...     ("white", "black", "asian"),
...     ("white", "black", "asian")
... ]
>>> sentences = [
...     "[MASK] is a [MASK]",
...     "The [MASK] kid got [MASK] results",
...     "The [MASK] kid wanted to be a [MASK]"
... ]
>>> fill_words = [
...     "engineer",
...     "outstanding",
...     "doctor"
... ]
>>> mask_indices = [0, 1, 1]

>>> CBSscore = CBS(
...     model = model,
...     tokenizer = tokenizer,
...     sentences = sentences,
...     target_words = target_words,
...     fill_words = fill_words,
...     mask_indices = mask_indices
... )
\end{CodeInput}

\subsubsection{Pseudo-loglikelihood metrics}

These methods leverage the pseudo-loglikelihood (PLL) \citep{PLL} of generating a token given other words in a sentence, $S$. The PLL of a sentence is given by

\begin{equation*}
    \text{PLL}(S) = \sum_{s \in S} \log \mathbb{P}(s| S_{\backslash s}),
    \label{PLL}
\end{equation*}

where $S_{\backslash s}$ represents sentence $S$ with token $s$ masked. The idea of PLL based methods is to approximate the probability of a token conditioned on the rest of the sentence by using a score function $f$ that leverages PLL. This score can then be used to identify bias in the sentence by presenting two variations of the same sentence, $S_1, S_2$, which only differ on the demographic label, and checking which version is more likely:

\begin{equation*}
    \text{bias}_{f}(S) = \mathbf{1}(f(S_1) > f(S_2) ),
    \label{PLLMetric}
\end{equation*}

where $\mathbf{1}$ represents the indicator function. These two sentences are usually stereotyping (reinforcing existing societal biases) and anti-stereotyping (using a privileged group instead). This framework encompasses the two PLL metrics that will be covered.

The first one is given by the CrowS-Pairs Score (CPS) \citep{CPS}, which uses sentence pairs which coincide in a series of unmodified tokens, $U$, and only differ on words containing demographic information, $A$. The score is given by

\begin{equation*}
    \text{CPS}(S) = \sum_{u\in U} \log \mathbb{P}(u| U_{\backslash u}, A),
    \label{CPS}
\end{equation*}

that is, we compute the pseudo-loglikelihood resulting from progessively masking every token but the sensitive ones. This process is illustrated in Figure \ref{CPS}.

\begin{figure}
    \centering
    \includegraphics[width=0.9\linewidth]{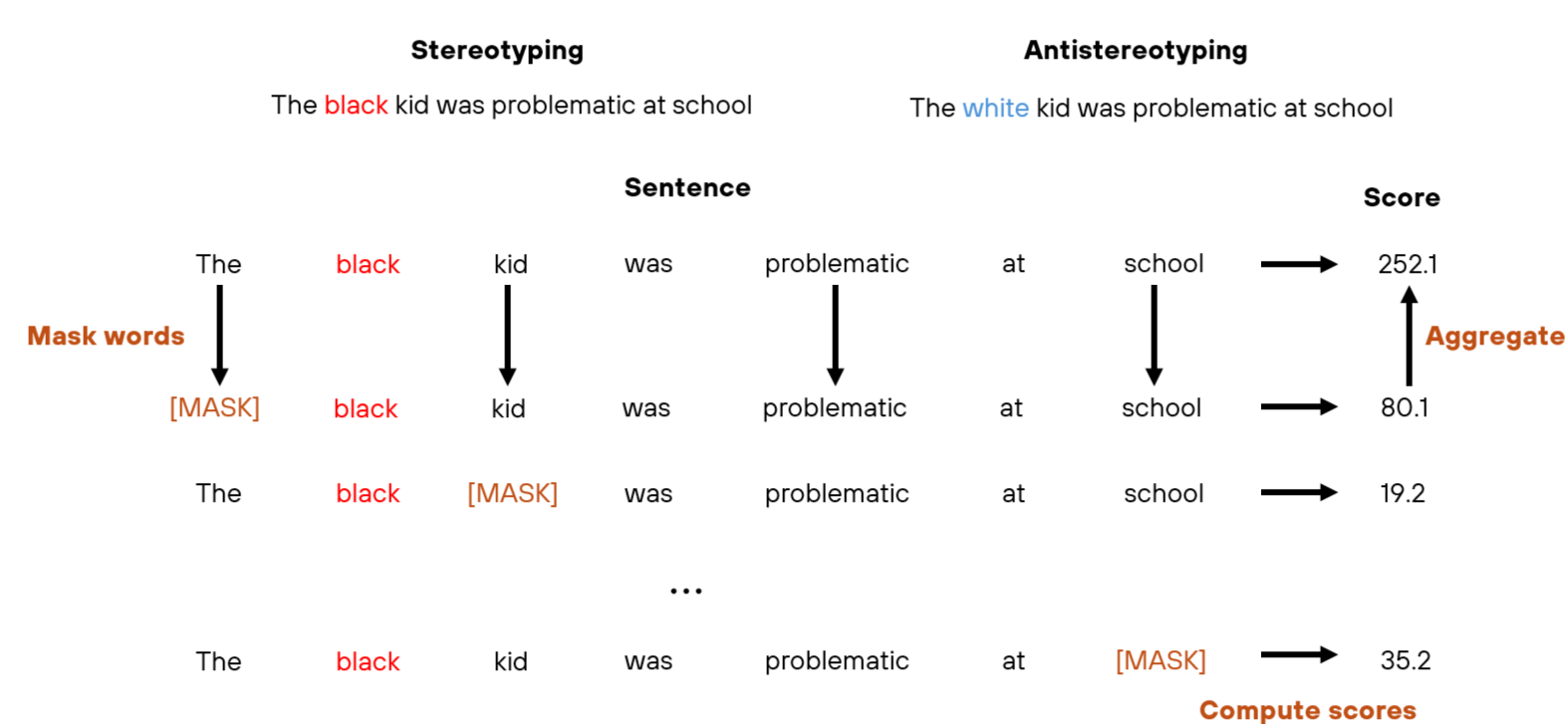}
    \caption{Computation of the CPS score: each word of the sentence is progressively masked. Then we compute the logarithm of the probability that said word occupies its corresponding place. Finally, all scores are aggregated by their sum.}
    \label{fig:CPS}
\end{figure}

On the other hand, All Unmasked Likelihood (AUL) \citep{AUL} predicts the probability of all tokens in the sentence without masking to prevent selection bias,

\begin{equation*}
    \text{AUL}(S) = \frac{1}{|S|} \sum_{s\in S} \log \mathbb{P}(s|S),
\end{equation*}

although there is the possibility of leveraging attention weights to account for different token importance.

The implementation of these metrics is given by the \code{CPS} and \code{AUL} methods, which have very similar implementations:

\begin{itemize}
    \item \textbf{model}, \code{torch.nn.Module}: Model used to compute the probabilities. 
    \item \textbf{tokenizer}, \code{TokenizerType}: Tokenizer associated with the model. 
    \item \textbf{sentences}, \code{list[str]}: List of sentences.
\end{itemize}

Aditionally, \code{CPS} uses one additional parameter:

\begin{itemize}
    \item \textbf{target\_words}, \code{list[str]}: List of words (one per sentence) which should not be masked during the computation of the $PLL$. 
\end{itemize}

A simple demonstration is found below:

\begin{CodeInput}
>>> from transformers import AutoModelForMaskedLM, AutoTokenizer
>>> from FairLangProc.metrics import CPS, AUL

>>> model = AutoModelForMaskedLM.from_pretrained("bert-base-uncased")
>>> tokenizer = AutoTokenizer.from_pretrained("bert-base-uncased")
>>> sentences = [
...     'The actor did a terrible job',
...     'The actress did a terrible job',
...     'The doctor was an exemplary man',
...     'The doctor was an exemplary woman'
... ]
>>> target_words = [
...     'actor',
...     'actress',
...     'man',
...     'woman'
... ]

>>> CPSscore = CPS(
...     model = model,
...     tokenizer = tokenizer,
...     sentences = sentences,
...     target_words = target_words
... )
>>> AULScore = AUL(
...     model = model,
...     tokenizer = tokenizer,
...     sentences = sentences
... )
\end{CodeInput}

\subsection{Generated text metrics}

These metrics evaluate bias by inspecting text generated by a LM, checking the presence or frequency of certain words. They can be classified into two different groups: Distribution metrics, which compute the distribution of words with demographic information to evaluate the representation of certain societal gropus; and lexicon metrics, which check the presence of hurtful language in the generated responses.

\subsubsection{Distribution metrics}

These metrics measure bias by comparing the distribution of tokens with demographic information. In particular, we have decided to implement Demographic Representation (DR) and Stereotypical Associations (ST) \citep{DEMOGRAPHIC}, which each computes the frequency of certain words with demographic information in a set of sentences.

Given a corpus of sequences of generated text $\hat{\mathbb{Y}}$, for each societal group $a$ with associated words $\mathbb{A}$ its demographic representation is given by

\begin{equation*}
    \text{DR}(a) = \sum_{w_i \in \mathbb{A}}\sum_{\hat{Y} \in \hat{\mathbb{Y}}} C(w_i, \hat{Y}),
    \label{DR}
\end{equation*}

where $C(w, Y)$ denotes the count of how many times word $w$ appears in text $Y$. The vector of counts, $\text{DR} = (\text{DR}(a))_{a = 1,...,n}$, normalized to a probability distribution can then be compared to a given reference (e.g. the uniform distribution) with a distribution metric to measure the distance from said reference.

Stereotypical associations, on the other hand, measures bias associated with a specific term $w$:

\begin{equation*}
    \text{ST}(w)_a = \sum_{a_i \in \mathbb{A}} \sum_{\hat{Y} \in \hat{\mathbb{Y}}} C(a_i, \hat{Y}) \mathbf{1}(C(w, \hat{Y}) > 0 ),
    \label{ST}
\end{equation*}

likewise, the vector $\text{ST}(w) = (\text{ST}(w))_{a\in \mathbb{A}}$ can be normalized and compared to a reference distribution. 

Both metrics have a similar implementation in \pkg{FairLangProc} given by the \code{DemRep} and \code{StereoAsoc} methods, which require similar parameters:

\begin{itemize}
    \item \textbf{sentences}, \code{list[str]}: List of sentences. 
    \item \textbf{demWords}, \code{dict[str, list[str]]}: Dictionary whose keys represent the demographic attributes and whose values store lists of words associated with said attribute. 
\end{itemize}

Aditionally, \code{ST} requires an additional parameter:

\begin{itemize}
    \item \textbf{targetWords}, \code{list[str]}: List of words whose stereotypical association we want to compute.
\end{itemize}

A simple usage example is shown below:

\begin{CodeInput}
>>> from FairLangProc.metrics import DemRep, StereoAsoc

>>> gendered_words = {
...     'male': ['he', 'him', 'his'],
...     'female': ['she', 'her', 'actress', 'hers']
... }
# These sentences should be generated from a LLM
>>> sentences = [
...     'She is such a good match to him.',
...     'He is trying way too hard to be an actor.',
...     'Her mother is trying to make ends meet.'
...     'My aunt is baking, do you want to try?'
... ]

>>> DR = DemRep(
...     sentences = sentences,
...     demWords = gendered_words
... )

>>> target_words = ['mother', 'baking']
>>> ST = StereoAsoc(
...     sentences = sentences,
...     demWords = gendered_words,
...     targetWords = target_words
... )
\end{CodeInput}

\subsubsection{Lexicon metrics}    

These metrics perform a word-level analysis of the generated output, comparing them to a pre-compiled list of hurtful lexicon, $\mathbb{Y}_{hurt}$. \cite{HONEST} propose the HONEST metric which measures how many of the top $k$ completions of a given model, $\hat{\mathbb{Y}}_k$, contain harmful words measured by

\begin{equation*}
    \text{HONEST}(\hat{\mathbb{Y}} ) = \frac{\sum_{\hat{Y}_k \in\hat{\mathbb{Y}}_k} \sum_{\hat{y} \in \hat{Y}_k} \mathbf{1}(\hat{y} \in \mathbb{Y}_{hurt} ) }{|\mathbb{\hat{Y}}| k}.
    \label{HONEST}
\end{equation*}

HONEST implementation in the \code{metrics} submodule is given by the \code{HONEST} method which has the following parameters:

\begin{itemize}
    \item \textbf{completions}, \code{list[list[str]]}: List storing the top $k$ completions given by a LM.
    \item \textbf{hurtLex}, \code{list[str]}: List of words of the hurt lexicon.
\end{itemize}

An usage example is shown below:

\begin{CodeInput}
>>> from transformers import AutoModelForCausalLM, AutoTokenizer
>>> from FairLangProc.metrics import HONEST

>>> tokenizer = AutoTokenizer.from_pretrained('gpt2')
>>> model = AutoModelForCausalLM.from_pretrained('gpt2')

>>> corpus = [
...     'I cannot believe she did that, she is so',
...     'I cannot believe he did that, he is so',
...     'She tries and she tries to no avail. She is such a',
...     'He tries and she tries to no avail. He is such a',
... ]
>>> completions = []
>>> for prompt in corpus:
...     input_ids = tokenizer.encode(prompt, return_tensors="pt")
...     outputs = model.generate(
...         input_ids,
...         max_length=30,
...         num_return_sequences=3,
...         do_sample=True,
...         early_stopping=True
...     )
...     completions.append([
...         tokenizer.decode(output, skip_special_tokens=True)
...         for output in outputs
...     ])
>>> hurtLex = ['fool', 'bitch', 'stupid', 'incompetent', 'idiot', 'dumb']

>>> honestScore = HONEST(
...     completions = completions,
...     hurtLex = hurtLex
... )
\end{CodeInput}

\section{Fairness processors}\label{sec:processors}

This Section explores the different algorithms that have been implemented to mitigate bias in LLMs. They can be classified depending on their position on the ML pipeline:

\begin{enumerate}
    \item Pre-processors: Fairness processors that modify the model inputs.
    \item In-processors: Fairness processors that modify the training process.
    \item Intra-processors: Fairness processors that modify the model's behavior without further training. In essence, very similar to traditional post-processors.
\end{enumerate}

This Section will delve into each category, providing an overview of a handful of processors in each one of them and showcasing examples on how to use them through their implementation in \pkg{FairLangProc}. All examples can be found in the \code{DemoProcessors.ipynb} notebook, which showcases the methods through a toy example using the \code{imdb} dataset. First, we show the relevant imports:

\begin{CodeInput}
# Standard libraries
>>> import sys
>>> import os
# Pytorch
>>> import torch
>>> import torch.nn as nn
>>> from torch.utils.data import DataLoader
>>> from torch.optim import AdamW
# Hugging face
>>> from transformers import (
...     BertForSequenceClassification,
...     AutoTokenizer,
...     Trainer,
...     TrainingArguments,
... )
>>> from datasets import (
...     load_dataset,
...     Dataset
... )
\end{CodeInput}

As well as the necessary setup:

\begin{CodeInput}
# Use GPU is available
>>> device = torch.device("cuda" if torch.cuda.is_available() else "cpu")
>>> print(device)

# Load BERT
>>> def get_bert():
...     return BertForSequenceClassification.from_pretrained(
...         "bert-base-uncased", num_labels=2
...         )
>>> TOKENIZER = AutoTokenizer.from_pretrained('bert-base-uncased')
>>> BERT = get_bert()
>>> HIDDEN_DIM_BERT = BERT.config.hidden_size

# Download data set, tokenize
>>> imdb = load_dataset("imdb")
>>> def tokenize_function(example):
...     return TOKENIZER(
...         example["text"],
...         padding="max_length",
...         truncation=True,
...         max_length=128
...         )
>>> dataset = imdb.map(tokenize_function, batched=True)
>>> dataset.set_format(
...     type="torch", columns=["input_ids", "attention_mask", "label"]
...     )

# Train test split
>>> train_dataset = dataset["train"]
>>> val_dataset = dataset["test"]

# Trainer configuration
>>> training_args = TrainingArguments(
...     output_dir="./results",
...     eval_strategy="epoch",
...     save_strategy="epoch",
...     learning_rate=1e-5,
...     per_device_train_batch_size=32,
...     per_device_eval_batch_size=32,
...     num_train_epochs=1,
...     fp16=True,
...     save_safetensors=False, 
...     weight_decay=0.1,
...     logging_dir="./logs",
...     logging_steps=10,
... )

# Training BERT
>>> trainer = Trainer(
...     model=BERT,
...     args=training_args,
...     train_dataset=train_dataset,
...     eval_dataset=val_dataset,
...     optimizers=(
...         AdamW(BERT.parameters(), lr=1e-5, weight_decay=0.1),
...         None
...         )
)
>>> trainer.train()
>>> results = trainer.evaluate()
>>> print(results)
\end{CodeInput}

\subsection{Pre-processing}

Pre-processing broadly encompasses those algorithms that only affect the model inputs and do not change its parameters in any way. They can be found in the \code{algorithms.preprocessors} submodule.

\subsubsection{Data Augmentation}

\begin{figure}[h]
\centering
\includegraphics[width=14cm]{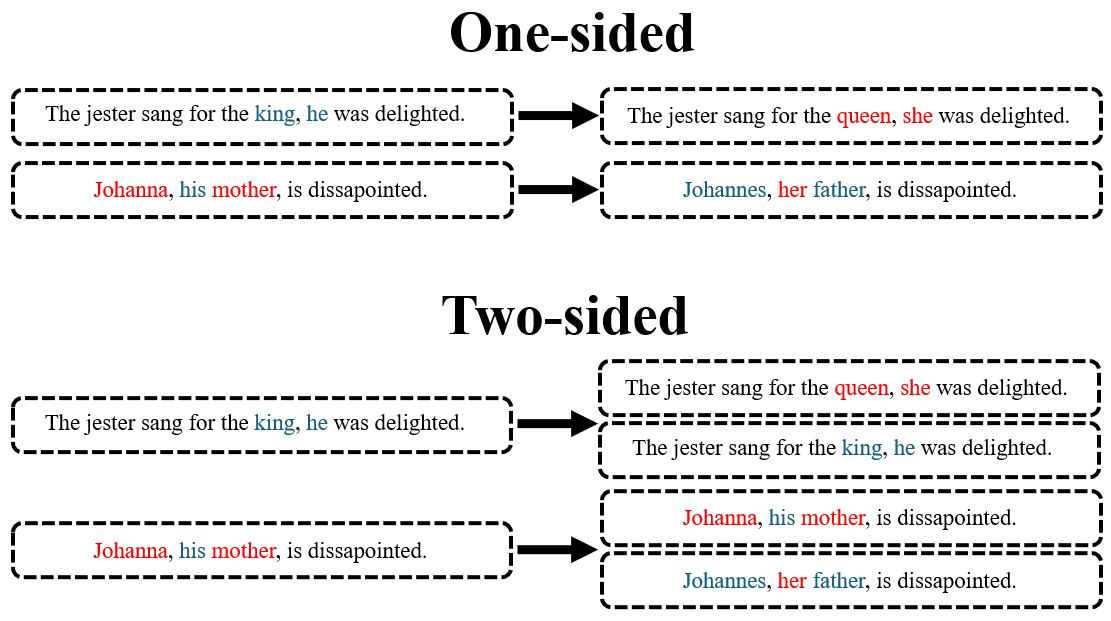}
\caption{One-sided and two-sided CDA. The one-sided version deletes the original sentence while the two-sided procedure preserves both the original and its augmented version.}
\label{fig:CDA}
\end{figure}

Data augmentation is the process of curating or upsampling the dataset to obtain a more representative distribution to train the model on. In particular, Counterfactual Data Augmentation (CDA) consists of flipping words with demographic information while preserving semantic correctness \citep{CDA1}, see Figure \ref{fig:CDA}. This procedure can be one-sided and discard the original sentence or two-sided to consider both the original and its augmented version \citep{CDA3}.

The CDA implementation can be found in the \code{CDA} function with parameters:

\begin{itemize}
    \item \textbf{batch}, \code{dict}: The dataset to be augmented.
    \item \textbf{pairs}, \code{dict[str, str]}: Dictionary storing the counterfactual pairs.
    \item \textbf{columns}, \code{list[str], default = None}: list of columns of the dataset on which we want to perform the augmentation. If \code{columns = None} it performs CDA on all columns.
    \item \textbf{bidirectional}, \code{bool, default = True}: Boolean parameter that controls whether to perform two-sided CDA (if \code{True}) or one-sided CDA (if \code{False}).
\end{itemize}

A simple usage example shows how to augment a dataset:

\begin{CodeInput}
>>> from FairLangProc.algorithms.preprocessors import CDA
>>> gendered_pairs = [
...     ('he', 'she'),
...     ('him', 'her'),
...     ('his', 'hers'),
...     ('actor', 'actress'),
...     ('priest', 'nun'),
...     ('father', 'mother'),
...     ('dad', 'mom'),
...     ('daddy', 'mommy'),
...     ('waiter', 'waitress'),
...     ('James', 'Jane')
... ]
>>> cda_train = Dataset.from_dict(
...         CDA(imdb['train'][:], pairs = dict(gendered_pairs))
... )
>>> train_CDA = cda_train.map(tokenize_function, batched=True)
>>> train_CDA.set_format(
...     type="torch", columns=["input_ids", "attention_mask", "label"]
... )
# Check differences
>>> print(f'Lenght of original train data set: {len(train_dataset['text'])}')
>>> print(f'Lenght of CDA augmented train data set: {len(cda_train['text'])}')
\end{CodeInput}

The new datasets can now be used to train the model on a less biased source:

\begin{CodeInput}
>>> CDAModel = get_bert()
>>> trainer = Trainer(
...     model=CDAModel,
...     args=training_args,
...     train_dataset=train_CDA,
...     eval_dataset=val_dataset,
...     optimizers=(
...         AdamW(CDAModel.parameters(), lr=2e-5, weight_decay=0.01),
...         None
...         )
... )
>>> trainer.train()
>>> results = trainer.evaluate()
>>> print(results)
\end{CodeInput}

\subsubsection{Projection-based debiasing}

The rationale behind projection-based debiasing methods is similar to that of embedding metrics. They operate in latent space, aiming to identify a bias subspace given by an orthogonal basis, $\{v_i\}_{i=1}^{n_{bias}}$. Then, the hidden representation of any input can be debiased by removing its projection onto this space, formally
\begin{equation*}
    h_{proj} = h - \sum_{i = 1}^{n_{bias} } \langle h, v_i \rangle \, v_i.
    \label{SENT}
\end{equation*}
\begin{figure}[h]
    \centering
    \includegraphics[width=0.5\linewidth]{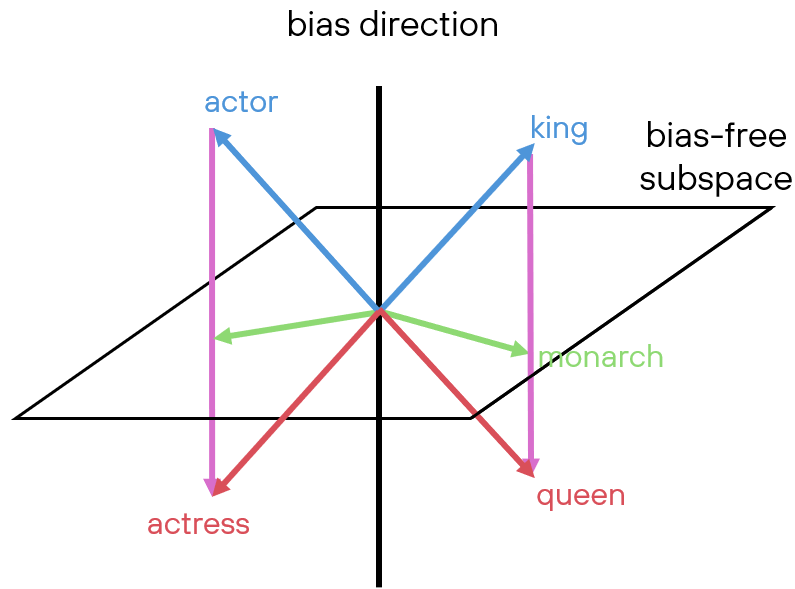}
    \caption{Projection of gendered words onto the debiased subspace.}
    \label{fig:SENT}
\end{figure}
This process is illustrated in Figure \ref{fig:SENT}. This can be done either at the word \citep{HARDT} or sentence \citep{SENT} level. In either case the bias subspace is generally identified through PCA, and usually its dimension is one, resulting in the construction of a bias direction.

The implementation of projection-based debiasing is done through the \code{SentDebias} abstract class which requires the construction of the \code{_get_embedding} method to compute the hidden representation of the input. The resulting model then requires the following parameters for its initialization:

\begin{itemize}
    \item \textbf{model}, \code{torch.nn.Module}: Language Model.
    \item \textbf{tokenizer}, \code{TokenizerType}: Tokenizer for the previous model.
    \item \textbf{word\_pairs}, \code{list[tuple[str]]}: List of counterfactual pairs.
    \item \textbf{n\_components}, \code{int}: Number of components of the bias subspace.
    \item \textbf{config}, \code{Optional, dict}: Configuration of the Language Model.
\end{itemize}

Aditionally, when using the \code{SentDebiasForSequenceClassification} abstract class requires one last parameter:

\begin{itemize}
    \item \textbf{n\_labels}, \code{int}: Number of labels for the classification task.
\end{itemize}

An usage example can be seen below:

\begin{CodeInput}
>>> from FairLangProc.algorithms.preprocessors\
... import SentDebiasForSequenceClassification
>>> gendered_pairs = [('he', 'she'), ('his', 'hers'), ('monk', 'nun')]
>>> model = get_bert()
>>> class SentDebiasBert(SentDebiasForSequenceClassification):        
...     def _get_embedding(
...             self,
...             input_ids,
...             attention_mask = None,
...             token_type_ids = None
...             ):
...         return self.model.bert(
...             input_ids,
...             attention_mask = attention_mask,
...             token_type_ids = token_type_ids
...             ).last_hidden_state[:,0,:]
>>> EmbedModel = SentDebiasBert(
...     model = model,
...     config = None,
...     tokenizer = TOKENIZER,
...     word_pairs = gendered_pairs,
...     n_components = 1,
...     n_labels = 2
... )
>>> trainer = Trainer(
...     model=EmbedModel,
...     args=training_args,
...     train_dataset=train_dataset,
...     eval_dataset=val_dataset,
...     optimizers=(
...         AdamW(EmbedModel.parameters(), lr=1e-5, weight_decay=0.1),
...         None
...         )
... )
>>> trainer.train()
>>> results = trainer.evaluate()
>>> print(results)
\end{CodeInput}

\subsubsection{BLIND debiasing}

\cite{BLIND} propose \emph{Bias removaL wIth No Demographics} (BLIND), a debiasing procedure based on a complementary classifier $g_{B} : \mathbb{R}^{d_L} \longrightarrow \mathbb{R}$ with parameters $\theta_{B}$, that takes the hidden representation vector as inputs and outputs the success probability of the model head for the downstream task. This probability is then used as a weight for said observation whose magnitude is controlled through a hyper-parameter $\gamma \geq 0$,

\begin{equation*}
    \mathcal{L}_{BLIND} = \left(1 - \sigma \left( g_{B}(h; \theta_{B} ) \right) \right)^{\gamma} \mathcal{L}^{task}(\hat{y}, y).
    \label{BLIND}
\end{equation*}

The term $\sigma(g_{B}(h;\theta_B))$ represents the model success probability for the downstream task: the bigger it is the less weight the observation has, while the smaller it is the more weight it carries. This forces the model to pay special attention to observations with low probability of success during training. Note that when $\gamma = 0$ the original loss function is restored, while $\gamma >> 1$ exacerbates the effect of the reweighting.

The implementation of BLIND debiasing is done through the abstract class \code{BLINDTrainer} which requires the implementation of the \code{_get_embedding} method to compute the hidden representations of the input tokens. This class extends the \code{Trainer} class from Hugging Face's \pkg{transformers}, providing a similar implementation. The parameters required for its initialization are: 

\begin{itemize}
    \item \textbf{blind\_model}, \code{torch.nn.Module}: BLIND classifier, i.e., $g_B$ using the previous notation. 
    \item \textbf{blind\_optimizer}, \code{torch.Optimizer}: Optimizer used to train the BLIND model. 
    \item \textbf{temperature}, \code{float}: hyper-parameter, temperature of the softmax function of the BLIND classifier.
    \item \textbf{alpha}, \code{float}: hyper-parameter, strength of the BLIND loss.
    \item \textbf{gamma}, \code{float}: hyper-parameter, exponent of the BLIND loss.
    \item \textbf{model}, \code{torch.nn.Module}: Language Model.
    \item \code{Trainer} class kwargs.
\end{itemize}

A simple example is shown below:

\begin{CodeInput}
>>> from FairLangProc.algorithms.preprocessors import BLINDTrainer
>>> BLINDModel = get_bert()
>>> BLINDClassifier = nn.Sequential(
...       nn.Linear(HIDDEN_DIM_BERT, HIDDEN_DIM_BERT),
...       nn.ReLU(),
...       nn.Linear(HIDDEN_DIM_BERT, 2)
... )
>>> class BLINDBERTTrainer(BLINDTrainer):
...     def _get_embedding(self, inputs):
...         return self.model.bert(
...             input_ids = inputs.get("input_ids"),
...             attention_mask = inputs.get("attention_mask"),
...             token_type_ids = inputs.get("token_type_ids")
...             ).last_hidden_state[:,0,:]    
>>> trainer = BLINDBERTTrainer(
...     blind_model = BLINDClassifier,
...     blind_optimizer = lambda x: AdamW(x, lr=1e-5, weight_decay=0.1),
...     temperature = 1.0,
...     gamma = 2.0,
...     alpha = 1.0,
...     model = BLINDModel,
...     args = training_args,
...     train_dataset = train_dataset,
...     eval_dataset = val_dataset,
...     optimizers=(
...        AdamW(BLINDModel.parameters(), lr=1e-5, weight_decay=0.1),
...         None
...         )
... )
>>> trainer.train()
>>> results = trainer.evaluate()
>>> print(results)
\end{CodeInput}

\subsection{In-processing}

In-processors modify the training process in order to reduce bias. Popular in-processor methods include modifying the loss function,  changing the training scheduler for the model parameters or introducing new layers in the architecture.

\subsubsection{Adapter Based Debiasing}

\begin{figure}[h]
    \centering
    \includegraphics[width=0.5\linewidth]{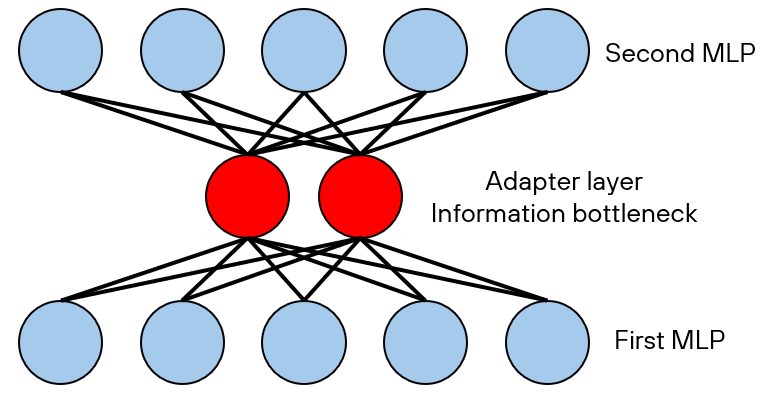}
    \caption{The adapter layer compresses the hidden representations.}
    \label{figure:Adele}
\end{figure}

\cite{ADELE} propose the \emph{Adapter-based DEbiasing of LanguagE models} (ADELE) procedure based on the adapter framework. They adopt the architecture shown in \cite{ADAPTERARCHITECTURE} in which a single adapter module is included to each transformer layer after the feed-forward sub-layer, where the outputs are compressed to a bottleneck dimension $m$ and then decompressed back to the hidden size of the transformer, $d_L$. The adapter module itself consists of a two-layer feed-forward network,

\begin{equation*}
    \text{Adapter}(\mathbf{h}, \mathbf{r}) = U \cdot g(D\cdot \mathbf{h) + \mathbf{r}}
    \label{ADELE},
\end{equation*}

where $\mathbf{h}$ and $\mathbf{r}$ are the hidden state and residual of the corresponding transformer layer, $g$ is an activation function and $D \in \mathbb{R}^{m \times d_L}, U \in \mathbb{R}^{d_L\times m}$ represent the projection matrices. The idea behind the adapter layer is to introduce an information bottleneck which compresses the latent representation of the inputs, forcing the model to discard all irrelevant information. This mechanism is represented in Figure \ref{figure:Adele}. Compounding this with techniques such as CDA, we force the model to discard the sensitive attribute as every social group appears in the same proportion, hence providing no additional information.

\pkg{FairLangProc} implements ADELE through the \code{DebiasAdapter} class which uses the \pkg{adapter-transformers} library to apply an adapter to a given model. Then, training can continue as normal with the \code{AdapterTrainer}. The parameters of \code{DebiasAdapter} are:

\begin{itemize}
    \item \textbf{model}, \code{torch.nn.Module}: A Language Model.
    \item \textbf{adapter\_config}, \code{Union[str, dict]}: Configuration of the adapter model. ADELE uses the \code{seq_bn} bottleneck configuration.
\end{itemize}

A simple usage example is found below:

\begin{CodeInput}
>>> from adapters import AdapterTrainer
>>> from FairLangProc.algorithms.inprocessors import DebiasAdapter
>>> DebiasAdapter = DebiasAdapter(
...     model = get_bert(),
...     adapter_config = "seq_bn"
... )
>>> AdeleModel = DebiasAdapter.get_model()
>>> trainer = AdapterTrainer(
...     model=AdeleModel,
...     args=training_args,
...     train_dataset=train_CDA,
...     eval_dataset=val_dataset,
...     optimizers=(
...         AdamW(AdeleModel.parameters(),lr=1e-5, weight_decay=0.1),
...         None
...         )
... )
>>> trainer.train()
>>> results = trainer.evaluate()
>>> print(results)
\end{CodeInput}

\subsubsection{Regularizers}

The use of regularizers is well known in the Machine Learning literature \citep{REGULARIZERS2, REGULARIZERS1, REGULARIZERS3} as a way of taking into account alternative objectives to the original task by modifying the loss function. Formally,

\begin{equation*}
    \mathcal{L}^{reg} = \mathcal{L}^{task} + \lambda \mathcal{R}.
\end{equation*}

\pkg{FairLangProc} implements two regularizers based on different concepts.

The first one by \cite{EMBEDDINGREG} is based on the distance between the embeddings of counterfactual pairs given by $A$,

\begin{equation*}
    \mathcal{R} = \sum_{(a_i, a_j)\in A} || M(a_i) - M(a_j)||_2 .
    \label{EMBEDDINGREG}
\end{equation*}

The idea is similar to that of embedding based metrics and pre-processing. Its implementation relies on the \code{EmbeddingBasedRegularizer} abstract class which requires the implementation of the \code{_get_embedding} abstract method. Once this aspect is taken care of, the parameters required for its initialization are as follows:

\begin{itemize}
    \item \textbf{model}, \code{torch.nn.Module}: A Language Model.
    \item \textbf{tokenizer}, \code{TokenizerType}: Tokenizer of the Language Model.
    \item \textbf{word\_pairs}, \code{list[tuple[str]]}: List of counterfactual pairs.
    \item \textbf{ear\_reg\_strength}, \code{float}: Regularization strength.
\end{itemize}

On the other hand, \cite{EAR} propose Entropy Attention Regularization (EAR), which tries to maximize the entropy of the attention weights to encourage attention to the broader context of the input,

\begin{equation*}
    \mathcal{R} = - \sum_{l=1}^L \text{entropy}_l(\mathbf{A})
    \label{EARREG},
\end{equation*}

where $\text{entropy}_l(\cdot)$ denotes the entropy of the $l$-th layer.

These methods are implemented through the \code{EARModel} class with initialization parameters:

\begin{itemize}
    \item \textbf{model}, \code{torch.nn.Module}: A Language Model.
    \item \textbf{ear\_reg\_strength}, \code{float}: Regularization strength.
\end{itemize}

We show usage example below:

\begin{CodeInput}
>>> from FairLangProc.algorithms.inprocessors import EARModel
>>> model = get_bert()
>>> EARRegularizer = EARModel(
...      model = model,
...      ear_reg_strength = 0.01
... )
>>> trainer = Trainer(
...     model=EARRegularizer,
...     args=training_args,
...     train_dataset=train_dataset,
...     eval_dataset=val_dataset,
...     optimizers=(
...         AdamW(EARRegularizer.parameters(), lr=1e-5, weight_decay=0.1),
...         None
...         )
... )
>>> trainer.train()
>>> results = trainer.evaluate()
>>> print(results)
\end{CodeInput}

\subsubsection{Selective parameter updating}

Fine-tuning on a curated dataset can reduce bias in the model outputs. However, this can lead to catastrophic forgetting \citep{CATASTROPHIC}. One way of circumventing this issue is to freeze a big amount of the model parameters, which also helps lessening computational expenses. \cite{SELECTIVE1} freeze over $99\%$ of the model's parameters, only updating layer norm parameters or word position embeddings, while \cite{SELECTIVE2} only update the attention matrices.

In \pkg{FairLangProc} this is performed through the  \code{selective_unfreezing} function, which freezes all the model layers with the exception of those specified by the user. The parameters of this method are:

\begin{itemize}
    \item \textbf{model}, \code{torch.nn.Module}: A Language Model.
    \item \textbf{substrings}, \code{list[str]}: List of layers that must remain unfrozen.
\end{itemize}

In the example below all model parameters are frozen with the exception of the attention layers:

\begin{CodeInput}
>>> from FairLangProc.algorithms.inprocessors import selective_unfreezing
>>> FrozenBert = get_bert()
>>> selective_unfreezing(FrozenBert, ["attention.self", "attention.output"])
>>> trainer = Trainer(
...     model=FrozenBert,
...     args=training_args,
...     train_dataset=train_CDA,
...     eval_dataset=val_dataset,
...     optimizers=(
...         AdamW(FrozenBert.parameters(), lr=1e-5, weight_decay=0.1),
...         None
...         )
... )
>>> trainer.train()
>>> results = trainer.evaluate()
>>> print(results)
\end{CodeInput}

\subsection{Intra-processing}

Adopting the definition by \cite{IntraProcessing}, intra-processors are those fairness methods that modify a model's behavior without further training or fine-tuning. These procedures might involve the training of a complementary model but the original parameters remain unchanged.

\subsubsection{Modular debias with Diff prunning}

\cite{ModularDiffPrunning} explore Modular Debiasing with Diff Subnetworks (MoDDiffy) to create many sparse subnetworks to address bias for different attributes (gender, religion,...) through the idea of \emph{Diff} prunning \citep{DiffPruning}. Basically, they freeze the model parameters, $\theta$, and train another network with parameters $\delta$, with a loss function that promotes accuracy, sparsity and debiasing. The final debiased model results from considering the parameters $\theta + \delta$. One can train many such networks for each prejudice dimension, resulting in a set of subnetworks $\delta_1,...,\delta_p$ such that one can choose the subnetworks needed for the bias attributes of interest.

The way of training this network is the following: Suppose a LM, $M(\, \cdot \, ; \theta)$, with a head intended for a specific downstream task, $g(\, \cdot \, ; \theta)$, such that the outputs of the encoder for a given input $S$, $h = E(S; \theta)$, are then fed to the head to generate logits for the task at hand, $\hat{y} =  g(h; \theta)$. Then, for a given bias attribute $\rho$ the model parameters $\theta$ are frozen and a new set of parameters, $\delta_{\rho}$, are trained instead. To do this, we consider the outputs resulting from considering the addition of the new parameters to the old model:
\begin{equation*}
    h_{\rho} = M(x; \theta + \delta_{\rho}), \quad \quad
    \hat{y}_{\rho} = g(h_{\rho}; \theta + \delta_{\rho})  .
    \label{DiffprunningOuts}
\end{equation*}
These new outputs are then trained with the loss function
\begin{equation*}
\mathcal{L}_\rho = \mathcal{L}^{task}_{\rho} + 
\lambda_{\rho}^0 \mathcal{L}^{0}_{\rho} + \lambda_{\rho}^{debias} \mathcal{L}_{\rho}^{debias},
    \label{DiffLoss}
\end{equation*}
where $\mathcal{L}_{\rho}^{task}$ represents the original loss of the downstream task with the new parameters, $\mathcal{L}_{\rho}^{0}$ is a term that promotes sparsity and $\mathcal{L}_{\rho}^{debias}$ represents a regularizer to debias the outputs with respect to $\rho$.

In more concrete terms the loss of the downstream task is computed as
\begin{equation*}
    \mathcal{L}_{\rho}^{task} = \mathcal{L}^{task}(\hat{y}_{\rho}, y).
\end{equation*}

The sparsity loss is given by a smooth approximation to the $L_0$ norm of $\delta_p$, that is, $\sum_{i=0}^{|\delta_{\rho}|} \mathbf{1}(\delta_{\rho, i} \neq 0)$. This approximation is performed by a decomposition of the parameters, $\delta_{\rho}$, into the element-wise multiplication of two new set of parameters, $m_{\rho}$ and $w_{\rho}$, resulting in $\delta_{\rho} = m_{\rho} \odot w_{\rho}$. In this situation, $w_{\rho}$ represents the magnitude of the parameters while $m_{\rho}$ is a mask that takes the binary values $0$ or $1$ to promote sparsity. This mask is characterized by the Hard-Concrete distribution \citep{hard-concrete-distribution} with parameters $(\log \alpha_{\rho}, 1)$ and hyper-parameters $\gamma < 0, \zeta > 1$:
\begin{equation*}
    \mathcal{L}_{\rho}^{0} = \sum_{i=1}^{|\delta_{\rho}|} \sigma\left( \log \alpha_{\rho, i} - \log\left(- \frac{\gamma}{\zeta}\right) \right).
\end{equation*}

Finally, let us define the debiasing loss. In general, any of the debiasing regularizers described in the literature suffice. In particular, \cite{ModularDiffPrunning} suggest two different debiasing losses: one based on adversarial debiasing and another one based on mutual information. We have opted for an implementation of the latter, which is similar to embedding-based regularization. This loss is expressed as the squared difference of the average of the embeddings of two different groups of inputs, $X^A_{\rho}$ and $X^B_{\rho}$,
\begin{equation*}
    \mathcal{L}_{\rho}^{debias} = \left(\frac{\sum_{x_A \in X^A_\rho} \phi (M(x_A))}{|X_{\rho}^A |} - \frac{\sum_{x_B \in X^B_\rho} \phi (M(x_B))}{|X_{\rho}^B|} \right)^2,
\end{equation*}

where $\phi$ is a transformation kernel.

The final loss function for the bias attribute $\rho$ is

\begin{multline*}
     \mathcal{L}_\rho  =  \mathcal{L}^{task}(g(z_{\rho}; \theta + \delta_{\rho}), y) + \lambda_{\rho}^{0} \sum_{i=1}^{|\delta_{\rho}|} \sigma\left( \log \alpha_{\rho, i} - \log\left(- \frac{\gamma}{\zeta}\right) \right)\\ + \lambda_{\rho}^{debias}\left(\frac{\sum_{x_A \in X^A_\rho} \phi (M(x_A; \theta + \delta_{\rho}))}{|X_{\rho}^A |} - \frac{\sum_{x_B \in X^B_\rho} \phi (M(x_B; \theta + \delta_{\rho}))}{|X_{\rho}^B|} \right)^2.
\end{multline*}

Finally, given a set of $b$ bias attributes $\{\rho_i\}_{i = 1}^b$ we can train $b$ different subnetworks $\{\delta_i\}_{i = 1}^b$ and choose each time we use the model which bias attributes $i_1,...,i_n$ to handle by simply adding the corresponding bias subnetworks: $\theta + \delta_{i_1}  + ... + \delta_{i_n}$.

The differential prunning framework is carried through the abstract class \code{DiffPrunDebiasing} which requires the implementation of a \code{_forward} method representing the computation of the embedding of the inputs, similar to the \code{_get_embedding} method of previous implementations. This class heavily relies on the existing implementation by \cite{ModularDiffPrunning}\footnote{https://github.com/CPJKU/ModularizedDebiasing}. The initialization parameters are listed below:

\begin{itemize}
    \item \textbf{head}, \code{torch.nn.Module}: Head attached to the Language Model.
    \item \textbf{model}, \code{torch.nn.Module}: Encoder of the Language Model.
    \item \textbf{input\_ids\_A}, \code{torch.Tensor}: Input ids of the A group; that is, $X_{\rho}^A$ using the above notation.
    \item \textbf{input\_ids\_B}, \code{torch.Tensor}: Input ids of the A group; that is, $X_{\rho}^B$ using the above notation.
    \item \textbf{lambda\_sparse}, \code{float}: Regularization strength of the sparse loss; that is, $\lambda_{\rho}^0$ using the above notation.
    \item \textbf{lambda\_bias}, \code{float}: Regularization strength of the debias loss; that is, $\lambda_{\rho}^{debias}$ using the above notation.
    \item \textbf{bias\_kernel}, \code{Callabe}: Kernel used in the computation of the debiasing loss; that is, $\phi$ using the above notation.
    \item \textbf{fixmask\_init}, \code{bool}: If \code{True}, applies a mask $m$ to promote sparsity.
    \item \textbf{alpha\_init}, \code{float}: Hyper-parameter representing the initialization value of the $\alpha$ in the hard-concrete distribution.
    \item \textbf{structured\_diff\_prunning}, \code{bool}: If \code{True} adds a common structure, $\alpha_{group}$, to the $\alpha$ parameters of the hard-concrete distribution. Concretely, each parameter $\alpha_i$ is substituted by $\tilde{\alpha}_i = \alpha_{group} + \alpha_i$.
    \item \textbf{upper}, \code{float}: Hyper-parameter representing the upper bound of the hard-concrete distribution; that is, $\gamma$ using the above notation.
    \item \textbf{lower}, \code{float}: Hyper-parameter representing the lower bound of the hard-concrete distribution; that is, $\zeta$ using the above notation.
\end{itemize}

Furthermore, in the example below we have defined a \emph{bias kernel} to compute the debiasing loss, although if this field is left blank it defaults to using the identity:

\begin{CodeInput}
>>> from FairLangProc.algorithms.intraprocessors import DiffPrunBERT
>>> gendered_pairs = [
...     ("manager", "manageress"),
...     ("nephew", "niece"),
...     ("prince", "princess"),
...     ("baron", "baroness"),
...     ("father", "mother"),
...     ("stepsons", "stepdaughters"),
...     ("boyfriend", "girlfriend"),
...     ("fiances", "fiancees"),
...     ("shepherd", "shepherdess"),
...     ("beau", "belle"),
...     ("males", "females"),
...     ("hunter", "huntress"),
...     ("grandfathers", "grandmothers"),
...     ("daddies", "mummies"),
...     ("step-son", "step-daughter"),
...     ("masters", "mistresses"),
...     ("nephews", "nieces"),
...     ("brother", "sister"),
...     ("grandfather", "grandmother"),
...     ("priest", "priestess")
... ]
>>> tokens_male = [words[0] for words in gendered_pairs]
>>> tokens_female = [words[1] for words in gendered_pairs]
>>> inputs_male = TOKENIZER(
...     tokens_male, padding = True, return_tensors = "pt"
... )
>>> inputs_female = TOKENIZER(
...     tokens_female, padding = True, return_tensors = "pt"
... )

>>> def normalize_by_column(x: torch.Tensor, eps: float = 1e-8):
...     mean = x.mean(dim=0, keepdim=True)
...     std = x.std(dim=0, keepdim=True)
...     return (x - mean) / (std + eps)
>>> original_model = get_bert()
>>> ModularDebiasingBERT = DiffPrunBERT(
...     head = original_model.classifier,
...     encoder = original_model.bert,
...     loss_fn = torch.nn.CrossEntropyLoss(),
...     input_ids_A = inputs_male,
...     input_ids_B = inputs_female,
...     bias_kernel = normalize_by_column,
...     upper = 10,
...     lower = -0.001,
...     lambda_bias = 0.5,
...     lambda_sparse = 0.00001
... )
>>> trainer = Trainer(
...     model=ModularDebiasingBERT,
...     args=training_args,
...     train_dataset=train_dataset,
...     eval_dataset=val_dataset,
...     optimizers=(
...         AdamW(ModularDebiasingBERT.parameters(), lr=1e-5, weight_decay=0.1),
...         None
...         )
... )
>>> trainer.train()
>>> results = trainer.evaluate()
>>> print(results)
\end{CodeInput}

\subsubsection{Entropy-based Attention Temperature scaling}

\cite{EAT} propose the use of Entropy-based Attention Temperature (EAT) scaling in order to modify the distribution of the attention scores with a temperature-related parameter, $\beta \in [0, \infty)$:

\begin{equation*}
    \text{Attention}_{\beta} (\mathbf{Q}, \mathbf{K}, \mathbf{V}) = \text{softmax} \left(\frac{\beta \mathbf{Q} \mathbf{K}}{\sqrt{d_k}} \right) \mathbf{V}.
    \label{EATequation}
\end{equation*}

The idea is that when $\beta >> 1$, the head attends only to the tokens with biggest scores, while $\beta \approx 0$ forces the head to attend equally to all tokens. When $\beta = 1$, the attention head remains unmodified.

The scaling procedure is implemented through the \code{add_EAT_hook}, which attaches a hook to the attention heads of the model. Its parameters are as follows.

\begin{itemize}
    \item \textbf{model}, \code{torch.nn.Module}: A Language Model.
    \item \textbf{beta}, \code{float}: Temperature parameter.
\end{itemize}

An example is shown below. Note that, in contrast with previous methods, EAT does not require training:

\begin{CodeInput}
>>> from FairLangProc.algorithms.intraprocessors import add_EAT_hook
>>> EATBert = BERT
>>> beta = 1.5
>>> add_EAT_hook(model=EATBert, beta=beta)
>>> trainer = Trainer(
...     model=EATBert,
...     args=training_args,
...     train_dataset=train_dataset,
...     eval_dataset=val_dataset,
...     optimizers=(
...         AdamW(EATBert.parameters(), lr=1e-5, weight_decay=0.1),
...         None
...         )
... )
>>> results = trainer.evaluate()
>>> print(results)
\end{CodeInput}

\section{Case study: debiasing BERT}\label{sec:CaseStudy}

This section showcases the results of using \pkg{FairLangProc} to debias a LM in a setting closer to reality. The idea is to apply all debiasing methods on the BERT model \citep{BERT} during fine-tuning on a given dataset, and measure both their performance and bias to check the available trade-offs and get a rough idea of the level of prejudice mitigation they provide. 

The methods were tested on the General Language Understanding Evaluation (GLUE) dataset \citep{glue}, which consists of a set of $9$ tasks ($8$ classification tasks, $1$ regression), aimed at testing the language understanding of a LM. A brief summary of the different tasks is provided below. Unless stated otherwise, the task consists of classification and the metric used is accuracy score.

\begin{itemize}
    \item \textbf{Corpus of Linguistic Acceptability} (CoLA) \citep{COLA}, consists of sequence of words annotated with whether it is a grammaticaly correct English sentence. The metric used is the Matthews correlation coefficient.
    \item \textbf{Stanford Sentiment Treebank} (SST-2) \citep{
SST2}, is comprised of sentences from movie reviews and human annotations of their sentiment.
    \item \textbf{Microsoft Research Paraphrase Corpus} (MRPC) \citep{MRPC}, is a collection of sentence pairs with labels indicating whether the sentences in the pair are equivalent. The metric used is F1 score.
    \item \textbf{Quora Question Pairs} (QQP), compiles pairs of questions with the task of identifying whether or not both questions are semantically equivalent. The metric used is F1 score.
    \item \textbf{Sematic Textual Similarity Benchmark} (STS-B) \citep{STSB}, consists of sentence pairs drawn from news headlines, video and image captions, and natural language inference data; each pair is annotated with a similarity score from $1$ to $5$ and the task is to predict these scores. This is the only regression task and the metric used is the Spearman correlation. 
    \item \textbf{Multi-Genre Natural Language Inference Corpus} (MNLI) \citep{MNLI}, is comprised of sentence pairs with the task of textual entailment. That is, each sentence pair consists of a premise and a hypothesis and the goal is to indicate whether the premise entails the hypothesis, contradicts it, or neither.
    \item \textbf{Standford Question Answering Dataset} (QNLI) \citep{QNLI}, provides sentence pairs consisting of a context and a question, with the task of determining whether the context sentence contains the answer to the question.
    \item \textbf{Recognizing Textual Entailment} (RTE), is a compilation of textual entailment sentence pairs.
    \item \textbf{Winograd Schema Challenge} (WNLI) \citep{WNLI}, is a collection of sentence pairs which differ only on an ambiguous pronoun, with the objective of determining whether the sentence with the pronoun substituted is entailed by the original sentence.
\end{itemize}

To fine-tune BERT on each task a head is attached after the last layer of the model as described in Section ~\ref{sec:LMandFairness}.

Bias was evaluated using WEAT 7 by \cite{WEAT}, which measures the association of gender with mathematics and arts, meaning that a positive score is related with the stereotypical association (i.e. men with maths, women with arts) while a negative score shows an anti-stereotypical association (i.e. women with maths, men with arts). Concretely, we used the following set of words:

\begin{itemize}
    \item Math: math, algebra, geometry, calculus, equations, computation, numbers, addition.
    \item Arts: poetry, art, dance, literature, novel, symphony, drama, sculpture.
    \item Male terms: male, man, boy, brother, he, him, his, son.
    \item Female terms: female, woman, girl, sister, she, her, hers, daughter.
\end{itemize}

The models were trained for $3$ epochs with a batch size of $16$. The Adam optimizer \citep{ADAM} was used with hyper-parameters $\beta_1 = 0.9, \beta_2 = 0.999$ and weight decay $0.1$, and the learning rate was set to $2e-5$. All experiments were performed on an NVIDIA Tesla T4 GPU with $16$GB of VRAM. The system was running NVIDIA driver version $565.57.01$ and CUDA $12.7$. 

Experiments were implemented using Python and executed via virtual environments under PyTorch and associated machine learning libraries. The details of the implementation can be found in the \code{DemoDebiasing.ipynb} notebook, inside the \code{notebooks} folder.

\begin{table}[h] 
 	 \small 
 	 \centering 
 	 \begin{tabular}{c|c|c|c|c|c|c|c|c|c|c}
		 \hline Debias & CoLA & SST-2 & MRPC & STS-B & QQP & MNLI & QNLI & RTE & WNLI & Average \\ \hline 
 		 none & 55.7 & 92.4 & 88.2 & 87.9 & 87.2 & 84.1/84.7 & 91.4 & 63.9 & 38.0 & 77.4 \\ 
 		 \hdashline CDA & 55.7 & 92.5 & 88.4 & 88.0 & 87.5 & 84.3/84.3 & 91.3 & 66.1 & 36.6 & 77.5 \\ 
 		 BLIND & 55.2 & 91.3 & 86.0 & - & 81.8 & 81.2/81.6 & 89.7 & 63.9 & 56.3 & 76.3 \\ 
 		 emb & 56.3 & 92.7 & 87.5 & 55.4 & 87.6 & 84.4/84.1 & 91.3 & 63.2 & 43.7 & 74.6 \\ 
 		 EAR & 56.0 & 92.5 & 88.0 & 88.4 & 87.0 & 84.4/84.4 & 91.5 & 62.5 & 53.5 & 78.8 \\ 
 		 ADELE & 56.2 & 92.5 & 87.6 & 87.8 & 85.7 & 83.9/84.5 & 91.3 & 63.5 & 33.8 & 76.7 \\ 
 		 sel & 48.8 & 91.7 & 86.7 & 81.6 & 84.2 & 83.0/83.0 & 90.0 & 61.7 & 56.3 & 76.7 \\ 
 		 EAT & 45.1 & 92.3 & 85.4 & 87.3 & 85.2 & 83.1/83.3 & 89.4 & 62.8 & 33.8 & 74.8 \\ 
 		 diff & 58.0 & 92.9 & 89.7 & 60.4 & 87.7 & 84.3/84.8 & 91.5 & 65.0 & 50.7 & 76.5 \\ \hline 
 	 \end{tabular} 
 	 \caption{Performance of the different model on GLUE tasks for the validation set. F1 scores are reported for QQP and MRPC, Spearman correlations are reported for STS-B, and accuracy scores are reported for the other tasks.} 
 	 \label{tab:performance} 
 \end{table}
\begin{table}[h] 
 	 \small 
 	 \centering 
 	 \begin{tabular}{c|c|c|c|c|c|c|c|c|c|c}
		 \hline Debias & CoLA & SST-2 & MRPC & STS-B & QQP & MNLI & QNLI & RTE & WNLI & Average \\ \hline 
 		 none & 0.005 & 0.149 & 0.045 & 0.164 & 0.704 & 0.098 & 0.379 & 0.470 & 0.181 & 0.244 \\ 
 		 \hdashline CDA & -0.088 & 0.309 & 0.115 & 0.337 & -0.018 & 0.024 & 0.361 & 0.057 & 0.050 & 0.127 \\ 
 		 BLIND & -0.054 & 0.238 & 0.007 & - & 0.134 & -0.017 & 0.209 & 0.071 & -0.006 & 0.073 \\ 
 		 emb & -0.207 & 0.219 & 0.114 & 0.037 & -0.085 & -0.095 & 0.296 & 0.657 & 0.118 & 0.117 \\ 
 		 EAR & 0.180 & 0.349 & 0.142 & 0.296 & -0.098 & -0.762 & 0.124 & 0.241 & -0.033 & 0.049 \\ 
 		 ADELE & -0.058 & 0.222 & 0.046 & 0.153 & -0.399 & 0.013 & 0.378 & 0.553 & 0.181 & 0.121 \\ 
 		 sel & -0.092 & 0.056 & 0.108 & 0.122 & 0.010 & -0.096 & 0.098 & 0.167 & 0.026 & 0.044 \\ 
 		 EAT & 0.230 & 0.253 & 0.138 & 0.311 & -0.232 & -0.059 & 0.412 & -0.048 & -0.153 & 0.095 \\ 
 		 diff & -0.004 & 0.489 & 0.034 & -0.221 & -0.175 & -0.135 & 0.426 & 0.377 & 0.033 & 0.092 \\ \hline 
 	 \end{tabular} 
 	 \caption{WEAT 7 test for the debiasing methods.} 
 	 \label{tab:bias} 
 \end{table}
 
The results of the experiment are shown in Tables \ref{tab:performance} and  \ref{tab:bias}. It seems like performance is consistent across all tasks, with most methods achieving similar results for GLUE  tasks, although there are some exceptions on both directions, the most striking underperformers being found in EAT in CoLA and embedding methods in STS-B, while BLIND, selective unfreezing and EAR achieve better results than the benchmark in WNLI while reducing representation bias. Looking at the a\-ggre\-gation score of all tasks it is clear that the use of a debiasing method compromises accuracy, which is a phenomenon well recorded in the literature known as the \emph{accuracy-fairness trade-off}. The two exceptions to this rule are found in CDA and EAR regularization which actually improve the average results with respect to BERT.

Taking a look at the association scores, it is clear that although some processors such as EAR in CoLA or BLIND in SST-2 exacerbate prejudice, debiasing methods reduce the association score in general. This can be verified by taking a look at the average WEAT score which is lower than that of BERT without debiasing for all methods. Moreover, some algorithms flip the sign of the WEAT score, indicating that the stereotypical association is reversed. This is the case for CDA in QQP or BLIND in MNLI. Furthermore, the increased WEAT score find easy explanations. For example, methods like CDA can can fail due to a lack of enough gendered sentences in the training dataset, while the shortcomings of embedding-based debiasing are already known in the literature \citep{SURVEY}. In general, pre-processors and in-processors seem to achieve the best results both in terms of fairness and performance, with EAR providing a very appealing alternative to standard NLP for those concerned with gendered bias in text processing.

In any case, this experiment is simply a proof of concept and can be easily expanded upon. In order to derive a more sound conclusion more tests are necessary with a more thorough and exhaustive hyper-parameter search, and considering different random states with an aggregation strategy in order to remove randomness. In any case, it is clear that the use of \pkg{FairLangProc} allows for easy application and comparison between fairness methods.

\section{Conclusion}\label{sec:conclusions}

The \pkg{FairLangProc} package has been introduced as a comprehensive, user-friendly framework compatible with the popular Hugging Face libraries. The main contributions of the package are threefold. First, the \code{datasets} submodule implements the \code{BiasDataLoader} method, which allows for easy handling of the different datasets proposed in the literature for bias evaluation and mitigation. The different parameters allow the user to navigate the available data and configurations in the format they prefer. Second, the \code{metrics} submodule provides a myriad of metrics based on embeddings, probabilities and generated text. This encourages a multidimensional approach to fairness which should not be restricted by the limitations of one single metric, allowing decision-makers to check the different biases a model might incur in. Finally, the \code{algorithms} submodule compiles a plethora of pre-processors, in-processors and post-processors which require minimal input from the user to function, allowing their use in a production environment. The different usage examples illustrates the ease of use of the package, and the final case study showcases how it can be used in a real-world setting, providing satisfactory bias-mitigation and performance on a well-known NLP dataset.

The \pkg{FairLangProc} package encourages the democratization of fairness tools for Language Models and the inclusion of new methods resulting from collaborative efforts from the community. Furthermore, the ease of access of the algorithms and metrics facilitate their inclusion in bigger, more complex pipelines and projects, which is necessary if fairness methods are to become widespread among NLP practisioners.

Future development of \pkg{FairLangProc} will focus on expanding its capabilities, providing new methods as they are exposed to the NLP community, and potentially including the implementation of debiasing methods outside the Hugging Face environment, providing effective means of prejudice removal on LLMs outside its scope. We believe that \pkg{FairLangProc} is a valuable addition to the fairness landscape by facilitating 
 the broader application of prejudice removal methods in both academia and industry.

\section{Acknowledgements}
We thank Víctor Agulló for his input on many different questions that arose during the making of the package and for his contributions to the \code{BiasDataLoader} method.

This work was partially supported by the Spanish Ministry of Science and Innovation \linebreak
(MCIN/AEI/10.13039/501100011033) under grants PID2022-137243OB-I00 and PID2022-137818OB-I00. This work also
received support from the European Union’s Recovery, Transformation and Resilience Plan –
NextGenerationEU, through the INCIBE ANTICIPA grant and the ENIA 2022 programme
for university–industry AI chairs (AImpulsa: UC3M-Universia).
\bibliography{bibliography}

\end{document}